\documentclass[a4paper]{article}

\usepackage{INTERSPEECH2020}
\usepackage{threeparttable} 
\usepackage{multirow,array}
\usepackage{listings}

\usepackage{xcolor}
\usepackage{soul}

\title{\textsc{deepSELF}: An Open Source Deep Self End-to-End Learning Framework}
\name{Tomoya Koike$^1$, Kun Qian$^{1*}$, Bj\"orn W.\ Schuller$^{2,3}$, Yoshiharu Yamamoto$^1$}
\address{
  $^1$Educational  Physiology  Laboratory,  The University of Tokyo, Japan\\
  $^2$GLAM -- Group on Language, Audio, \& Music, Imperial College London, UK\\
  $^3$Chair of Embedded Intelligence for Health Care and Wellbeing, University of Augsburg, Germany
  \thanks{Kun Qian is the \emph{Corresponding Author}.}}
\email{\{tommy, qian, yamamoto\}@p.u-tokyo.ac.jp, schuller@ieee.org}

\begin{document}

\maketitle
\begin{abstract}
We introduce an open-source toolkit, i.\,e., the deep Self End-to-end Learning Framework (\textsc{deepSELF}), as a toolkit of deep self end-to-end learning framework for multi-modal signals. To the best of our knowledge, it is the first public toolkit assembling a series of state-of-the-art deep learning technologies. Highlights of the proposed \textsc{deepSELF} toolkit include: First, it can be used to analyse a variety of multi-modal signals, including images, audio, and single or multi-channel sensor data. Second, we provide multiple options for pre-processing, e.\,g., filtering, or spectrum image generation by Fourier or wavelet transformation. Third, plenty of topologies in terms of NN, 1D/2D/3D CNN, and RNN/LSTM/GRU can be customised and a series of pretrained 2D CNN models, e.\,g., AlexNet, VGGNet, ResNet can be used easily. Last but not least, above these features, \textsc{deepSELF} can be flexibly used not only as a single model but also as a fusion of such.
\end{abstract}


\section{Introduction}
\label{sec_intro}

Feature engineering, as an essential and inevitable step in the \emph{classic} machine learning (ML) paradigm, has been studied in a plethora of works. On the one hand, well-designed features by employing expert domain knowledge (e.\,g., speech, image, or such as medicine) lead to efficient and robust ML models for individual and specific tasks. On the other hand, hand-crafted features by humans are expensive, time-consuming, and inflexible to create for practical artificial intelligence (AI) projects. \emph{Deep learning} (DL)~\cite{lecun2015deep}, a sub-discipline of ML, has increasingly evolved in comparison to the traditional paradigms in ML in recent decades. By using a series of non-linear transformations of the inputs, DL models can learn higher representations from raw data directly without any domain knowledge of a human expert. Benefiting from this revolution, researchers can free time focusing on exploring AI-supported scientific research rather than overcoming the barriers between their subjects and recent machine learning.

Among the state-of-the-art technologies, there are two popular methods, i.\,e., \emph{deep spectrum transfer learning} and \emph{deep end-to-end learning}. The former involves a generation of spectrum images and pretrained deep neural network models to extract higher representations. The latter is aimed to learn the higher representations directly from the raw signals by combining convolutional neural networks (CNN) and recurrent neural networks (RNN). 
One toolkit that supports \emph{deep spectrum transfer learning} is \textsc{DeepSpectrum}~\cite{amiriparian2017snore}, which enables users to extract features from a 2-dimensional spectrogram by a pretrained CNN.
It provides the tool for feature extraction and lets users select classifiers, rendering it impossible to fine-tune from pretrained CNN weights.
Another toolkit, named \textsc{End2You}~\cite{tzirakis2018end2you}, supports deep end-to-end learning with a fixed deep learning model structure. Assuming that a different model structure fits different data, the given inflexibility of model structures limits the range of applications of the toolkit.

Based on this background of limited effectiveness of fine-tuning from pretrained weights and the further disadvantages of existing toolkits, we developed the toolkit named \textsc{deepSELF}, which supports fine-tuning with pretrained CNNs and deep end-to-end learning. 
It accepts multi-modal inputs and multi-channel inputs. 
A variety of pretrained CNNs are provided in this toolkit, e.\,g.,  AlexNet~\cite{krizhevsky2012imagenet}, GoogLeNet~\cite{szegedy2015going}, VGG~\cite{simonyan2015}, ResNet~\cite{he2016deep}, ResNeXt~\cite{xie2017aggregated}, MobileNet~\cite{howard2017mobilenets}, WideResNet\cite{zagoruyko2016wide}, and ResNeXt-WSL~\cite{mahajan2018exploring}.
Furthermore, users of \textsc{deepSELF} can change not only hyperparameters such as learning rate and batch size, but also the structure of deep learning models such as the number of layers and hidden nodes, which enhances the model flexibility and leads to higher accuracy.

The remainder of this paper will be organised as follows: We will give an overview on  \textsc{deepSELF} in Section~\ref{sec_overview}. Then, experiments will be detailed out in Section~\ref{sec_exp} before a conclusion which is made in Section~\ref{sec_con}.

\section{System Overview}
\label{sec_overview}

\begin{figure*}[t!]
\centering
\includegraphics[width=16cm]{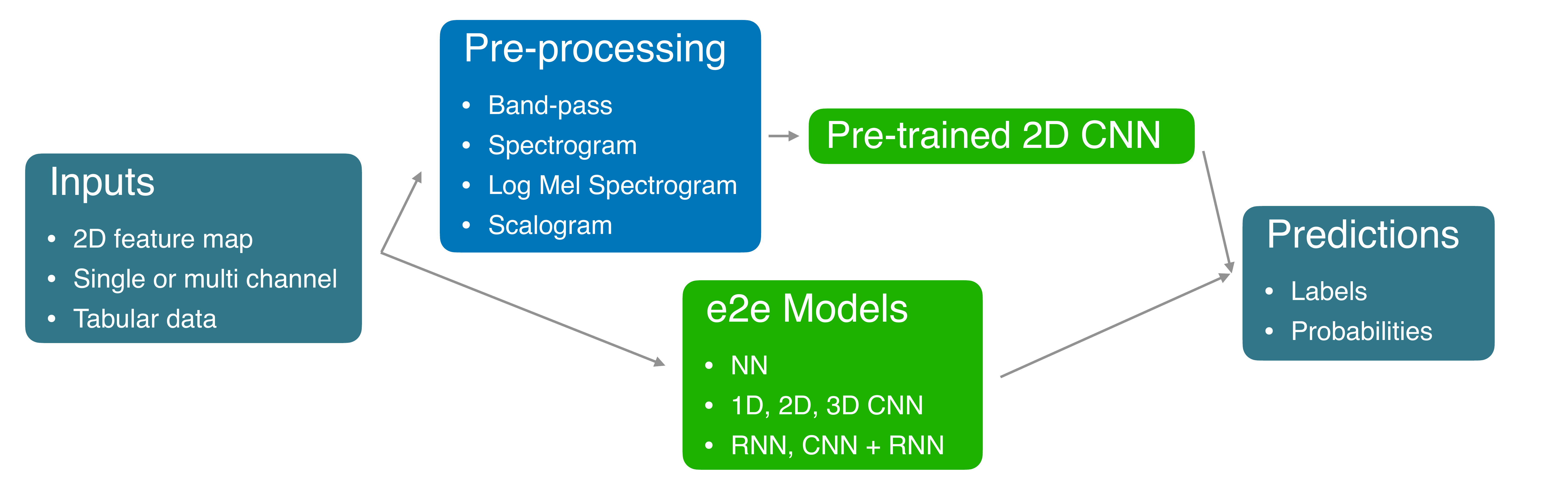}
\caption{An overview scheme of the \textsc{deepSELF} toolkit.}
\label{system-overview}
\end{figure*}

There are mainly four parts of the processing of \textsc{deepSELF}: Inputs, Pre-processing, Models,  and Predictions, as shown in Figure~\ref{system-overview} in detail.
As an input, \textsc{deepSELF} accepts 1-dimensional and 2-dimensional data, either from a single channel or a multi-channel data source. In the pre-processing part, band-pass filtering by a  fourth-order Butterworth filter can be used to cut off some frequencies from a raw signal. Following band-pass filtering, a 2-dimensional feature map can be calculated by one of the three methods: spectrogram, Log-Mel spectrogram, and scalogram. 

As a model to predict labels from inputs, \textsc{deepSELF} covers mainly four types of neural network structures: fully connected neural networks (FNN), convolutional neural networks (CNN), recurrent neural networks (RNN) and pretrained 2-dimensional CNN. CNN and RNN can be stacked in the direction from input to output, with automatically reshaping the feature map and keeping the axis of time steps. The hyperparameters that users can change are listed in Table~\ref{tab_hyperpara}. The parameters of training common in all neural networks are listed as general hyperparameters. Batch size is the number of instances in one batch of training, and the optimiser can be chosen from stochastic gradient descent (SGD) and Adam~\cite{kingma2015adam}. Both FNN and RNN have the number of hidden layers and hidden nodes as their structure parameters ranging in non-negative integer. As for the CNN structure parameters, \textsc{deepSELF} internally calculates the size of the output feature map in each hidden layer according to the user-determined parameters: the number of channels, kernel size, stride size, and padding size. For RNN models, conventional `normal' RNN, long short-term memory (LSTM)~\cite{hochreiter1997long} RNN, and gated recurrent unit (GRU)~\cite{chung2014empirical} RNN are the supported options. 
Bi-directional RNN are further supported, given that if non-causal handling of input data is possible, they usually proof effective for time-series input~\cite{amodei2015deep} with past and future temporal context. 
As an output, \textsc{deepSELF} can provide both predicted labels and probabilities from the best-scored model on a development dataset.

\subsection{Pretrained CNN}
\label{sec_pretrained}
There are many pretrained models publicly available, which are trained with a huge number of image datasets like ImageNet~\cite{krizhevsky2012imagenet}: VGG~\cite{simonyan2015}, MobileNet~\cite{howard2017mobilenets}, ResNet~\cite{he2016deep}, and ResNeXt~\cite{xie2017aggregated}. In the ImageNet 2014 challenge, the model known as VGG reached the second place in the classification track, which is characterised by 3$\times$3 convolutional filters and 16~\textasciitilde~19 convolutional layers, which is deeper than the previous CNN models at that time~\cite{simonyan2015}. ResNet~\cite{he2016deep} was introduced featuring residual blocks, which were proposed on the background that deeper CNN often achieve higher performance while it is challenging to train them due to the vanishing gradient~\cite{he2016deep}. Residual blocks are composed of two pathways from input to output of residual blocks, which are the normal convolutional layer and shortcut connections between input and output. The shortcut path does not affect the number of parameters and matrix calculation while making backpropagation easier because of the shorter path at the same time. In the ResNets, convolutions with 3$\times$3 filters cost heavily on computing. To reduce the computational cost, MobileNet V1 factorised a standard convolution into two convolutions: depthwise convolution and pointwise convolution by 1$\times$1 filters~\cite{howard2017mobilenets}. Inverted residual blocks, known as bottleneck layers, were added in MobileNet V2~\cite{sandler2018mobilenetv2}, reducing the number of parameters in comparison to MobileNet V1. To obtain better ability of image representation than ResNet, ResNeXt~\cite{xie2017aggregated} added group convolutions and reduced channel-wise compression rate, keeping the number of parameters the same with ResNet. Recently, weakly supervised learning with ResNeXt was proposed in~\cite{mahajan2018exploring}. It is pretrained in a weakly-supervised fashion on 940 million public images with 1.5\,k hashtags matching with 1\,000 ImageNet1K synsets, followed by fine-tuning on the ImageNet1K dataset.

\begin{table}[t]
\caption{Model parameters users can change from the execution command}
\label{tab_hyperpara}
\centering
\begin{threeparttable}
\begin{tabular}{ll}
\toprule
\textbf{Model}                & \textbf{Hyperparameters} \\ 
\midrule
\multirow{2}{*}{General}           & \emph{learning rate}; \emph{batch size}; \emph{\# of epochs}; \\
                                & \emph{optimiser}:~`sgd', `adam';. \\
\midrule
\multirow{2}{*}{NN}           & \emph{\# of hidden layers}:~1, 2, $\cdots$; \\
                              & \emph{\# of hidden nodes}:~1, 2, $\cdots$; \\
\midrule
\multirow{2}{*}{CNN}          & \emph{\# of channels}; \emph{kernel size}; \emph{stride size}; \\ 
                                & \emph{padding size}; \\
\midrule
\multirow{3}{*}{RNN}          & \emph{rnn type}:~`rnn', `lstm', `gru'; \\
                                & \emph{direction}:~`uni-directional', `bi-directional'; \\
                              & \emph{\# of hidden layers}; \emph{\# of hidden nodes}; \\
\midrule
\multirow{3}{*}{Pretrained}  & \emph{2D CNN}:~`alexnet', `googlenet', `vgg', \\
                                & `resnet', `resnext', `mobilenet', \\
                                & `wideresnet', `resnext-wsl'; \\
\bottomrule
\end{tabular}
\end{threeparttable}
\end{table}

\section{Experiments}
\label{sec_exp}

In this section, we report the effectiveness and usability of \textsc{deepSELF} with three kinds of input modalities. The effect of pre-training and fine-tuning will be seen by an experiment on an  image dataset. The comparison with other toolkits is conducted with a publicly available audio dataset, provided by~\cite{piczak2015esc}. To demonstrate the portability of different end-to-end deep learning structures, we provide another experiment executed on an EEG dataset. All of the experiments are evaluated with unweighted average recall~(UAR), which avoids over-optimistic conclusions for imbalanced distributions of labels among the test instances~\cite{schuller2009interspeech}.
Supported deep neural networks and pretrained CNNs are provided with the  \textsc{PyTorch}~\cite{steiner2019pytorch} backbone, version 1.4.0. \textsc{deepSELF} is written in and tested with \textsc{Python} 3.7, with a GPU and CUDA library version 10.1.

\subsection{Image data}
To demonstrate the usability of pretrained models with \textsc{deepSELF}, we use a facial expression dataset, which is publicly available from Kaggle~\cite{kaggle-face}. 
Labels are composed of six expressions: Anger, disgust, fear, happy, sad, surprise and neutral. 
We compared VGG16, ResNet, ResNeXt, and pretrained models of these with UAR in the test set as shown in Table~\ref{face-results}.
We can say that pre-training is effective to distinguish facial expressions, increasing the UAR by 11.9\,\% on average for the three models.

\begin{table}[t]
\centering
\caption{The results of VGG, ResNet, and ResNeXt on the facial expression classification task.}
\label{face-results}
\begin{threeparttable}
\begin{tabular}{lrr}
\toprule
\multicolumn{3}{c}{Dev UAR$\left[\,\% \,\right]$} \\ \hline
        & \multicolumn{2}{c}{Pretrained}                      \\ \hline
Model   & \multicolumn{1}{c}{False} & \multicolumn{1}{c}{True} \\ \hline
VGG   & 51.5                      & 60.0                     \\
ResNet  & 47.1                      & 58.2                     \\
ResNeXt & 44.1                      & \textbf{60.2}                     \\ 
\bottomrule
\end{tabular}
\end{threeparttable}
\end{table}

\subsection{Audio data}
ESC-10 is a labelled set of 400 environmental recordings equally from 10 classes, which means 40 clips per class. The clips' audio length is 5 seconds, sampled with 44.1\,kHz with a single channel.
We compare the results of a pretrained 2D CNN supported by \textsc{deepSELF} with the results of other models, as shown in Table~\ref{esc-results}.
The UAR was calculated by the average of a 5-fold cross-validation, noting that fold allocation of each instance is given by the database distributor 
and hence is strictly comparable.
ResNet with ImageNet pretrained (\textsc{deepSELF} ResNet in Table~\ref{esc-results}) scored 75.0\,\%UAR, while PANN pretrained on AudioSet  (\textsc{deepSELF} PANN in Table~\ref{esc-results}) scored 90.0\,\% which lies in the second place following Aytar et al.~\cite{aytar2016soundnet}.

\begin{table}[t]
\centering
\caption{The results of \textsc{deepSELF} and other models on the ESC-10 dataset}
\label{esc-results}
\begin{tabular}{lr}
\hline
Model           & \multicolumn{1}{l}{UAR {[}\%{]}} \\ \hline
Baseline~\cite{piczak2015esc}        & 72.7                             \\
\textsc{auDeep}~\cite{freitag2017audeep}       & 82.7                             \\
Piczak~\cite{piczak2015environmental}          & 80.3                             \\
Aytar et al.~\cite{aytar2016soundnet}   & \textbf{92.2}                             \\
\textsc{deepSELF} ResNet & 75.0                             \\
\textsc{deepSELF} PANNs  & \textbf{90.0}                             \\ \hline
\end{tabular}
\end{table}

\subsection{EEG data}
EEG data for automatic seizure detection is provided by the University of Bonn~\cite{andrzejak2001indications}. This database contains 5 sets of 100 records each. Set A to Set D are records without seizures from both healthy individuals and epilepsy patients. Set E are the records during seizures of epilepsy patients. The sampling rate of the data is 173.61\,Hz.
We tested 1D CNN, 1D CNN + RNN, and RNN with this data, as shown in Table~\ref{eeg-results}.
The 1D CNN is composed of 4 convolutional layers and 2 fully connected layers, while the RNN has 2 bi-directional layers of GRU and 1 fully connected layer. The stack of 1D CNN and RNN denoted as 1D CNN + RNN (GRU) in the Table~\ref{eeg-results} has 4 convolutional layers, 2 layers of bi-directional GRU, and 1 fully connected layer.

\begin{table}[t]
\centering
\caption{The results of DNN models of \textsc{deepSELF} on the EEG dataset}
\label{eeg-results}
\begin{tabular}{lrr}
\hline
Model     & \multicolumn{1}{l}{Dev UAR {[}\%{]}} & \multicolumn{1}{l}{Test UAR {[}\%{]}} \\ \hline
1D CNN       & 50.0                                 & 50.0                                  \\
RNN (GRU)       & 66.9                                 & 65.0                                  \\
1D CNN + RNN (GRU) & 71.2                                 & 40.0                                  \\ \hline
\end{tabular}
\end{table}

\section{Conclusion}
\label{sec_con}

We proposed an open-source toolkit named \textsc{deepSELF}, which can be flexibly implemented for classification tasks of mono- and multi-modal signals. The experiments shown were validated on  publicly accessible databases, including image, audio, and physiological signals (EEG in this study). We believe that \textsc{deepSELF} can be useful and efficient for many other machine learning tasks without using domain knowledge of a human expert. In future work, we will develop more features in \textsc{deepSELF} like federated learning~\cite{konevcny2016federated}, and evolving learning~\cite{zhang2018evolving}.

\section{Acknowledgements}

This work was partially supported by the Zhejiang Lab's International Talent Fund for Young Professionals (Project HANAMI), P.\,R.\,China, the JSPS Postdoctoral Fellowship for Research in Japan (ID No.\,P19081) from the Japan Society for the Promotion of Science (JSPS), Japan, and the Grants-in-Aid for Scientific Research (No.\,19F19081 and No.\,17H00878) from the Ministry of Education, Culture, Sports, Science and Technology (MEXT), Japan.

\bibliographystyle{IEEEtran}
\bibliography{references}


\end{document}